\pdfoutput=1

\documentclass[11pt]{article}

\usepackage{ACL2023}

\usepackage{times}
\usepackage{latexsym}

\usepackage[T1]{fontenc}

\usepackage[utf8]{inputenc}

\usepackage{microtype}

\usepackage{inconsolata}

\usepackage{graphicx}
\usepackage{amsmath}

\title{Teaching the Pre-trained Model to Generate Simple Texts for Text Simplification}

\author{$^1$Renliang Sun, $^2$Wei Xu, $^1$Xiaojun Wan \\
  $^1$Wangxuan Institute of Computer Technology, Peking University \\
  $^1$Center for Data Science, Peking University \\
  $^1$The MOE Key Laboratory of Computational Linguistics, Peking University \\
  $^2$School of Interactive Computing, Georgia Institute of Technology \\
  \texttt{sunrenliang@stu.pku.edu.cn} \quad \texttt{wei.xu@cc.gatech.edu} \\ \texttt{wanxiaojun@pku.edu.cn} \\}

\begin{document}
\maketitle
\begin{abstract}

Randomly masking text spans in ordinary texts in the pre-training stage hardly allows models to acquire the ability to generate simple texts. It can hurt the performance of pre-trained models on text simplification tasks.
In this paper, we propose a new continued pre-training strategy to teach the pre-trained model to generate simple texts. We continue pre-training BART, a representative model, to obtain SimpleBART. It consistently and significantly improves the results on lexical simplification, sentence simplification, and document-level simplification tasks over BART. At the end, we compare SimpleBART with several representative large language models (LLMs).

\end{abstract}

\section{Introduction}

Text simplification (TS) is a task in the field of natural language generation. It aims at rewriting a complex text into simple text while keeping the primary meaning intact \cite{laban2021keep}.

Recently, several works have leveraged pre-trained models for TS \cite{omelianchuk2021text, devaraj2022evaluating}.
However, problems arise when pre-trained models are applied to TS directly. In the pre-training stage, the model hardly acquires the ability to generate simple texts. The improvement of results on simplification tasks relies almost on the fine-tuning stage. It can hurt the performance of pre-trained models, especially for low-resource sub-tasks like lexical simplification. One reason for this shortcoming is the pre-training strategy. It randomly masks text spans in ordinary texts, teaching the model to generate ordinary texts rather than simple texts.

We are committed to adapting the pre-trained model to TS in this paper. The pre-trained model has gained the ability to generate ordinary texts, and it is costly to start pre-training from scratch. Therefore, we focus on the continued pre-training strategy \cite{gururangan2020don}.
We first aim to continue pre-training on simple texts because it contains plenty of simple words. In TS, simple texts are derived almost from SimpleWiki \cite{zhang2017sentence} and Newsela \cite{xu2015problems}.
We identify simple text spans in simple texts and dynamically replace them with <mask> tokens. 
Then, the pre-trained model will learn by reconstructing simple words. Meanwhile, we expect the pre-trained model to learn from ordinary texts. 
We use a dictionary to replace complex words in ordinary texts with simple words. We also ensure the quality of the replaced sentences.

Based on BART \cite{lewis2020bart}, we continue pre-training to teach it to generate simple texts and obtain SimpleBART. We then conduct experiments on three main tasks of TS: sentence simplification, lexical simplification, and document-level simplification. SimpleBART achieves consistent and noticeable improvements across several datasets on all three tasks over BART and several other baselines. The results illustrate that our proposed strategy helps the pre-trained model to gain the ability to generate simple texts.


\begin{figure*}[h]
\centering
\includegraphics[width=\textwidth]{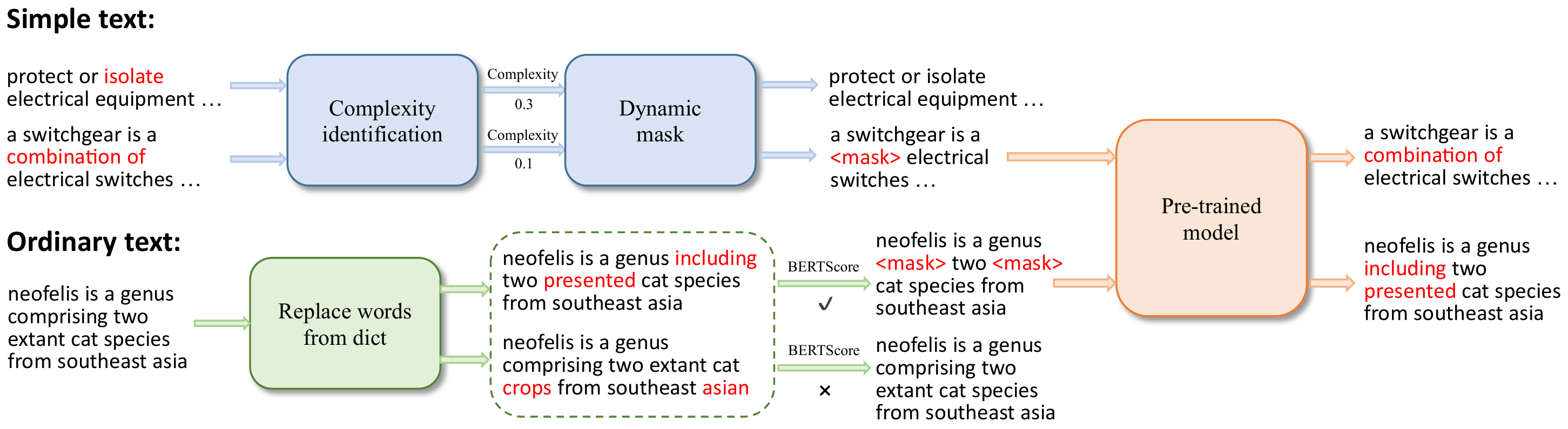} 
\caption{Overview of our continued pre-training strategy to teach the pre-trained model to generate simple texts.}
\label{fig:flowchart}
\end{figure*}

To summarize, our contributions include: (1) We propose a new continued pre-training strategy to teach the pre-trained model to generate simple texts. (2) We continue pre-training BART, a representative seq2seq model, to obtain SimpleBART. It can be used for several simplification tasks and achieve consistent performance improvement.  Code and SimpleBART will be released at \url{https://github.com/RLSNLP/SimpleBART}.

\section{Methodology}

As illustrated in Figure \ref{fig:flowchart}, our strategy is divided into two parts: learning dynamically to reconstruct simple words from simple texts and from ordinary texts where complex words are replaced with simple ones.

\subsection{Masking Simple Words in Simple Texts}
\label{sec:mask-simple}

We need to identify the simple words in simple texts at first. We take advantage of the DeepBlueAI model \cite{pan2021deepblueai} that achieves state-of-the-art results on the lexical complexity prediction task \cite{shardlow2021semeval}. A text span of length $n$ consists of $n$ words. The input to the DeepBlueAI model is a text span and the output is a complex value between 0 and 1.
The closer this value is to 0, the simpler the text span.


Unlike the previous constant mask probability, in our strategy, the simpler a text span is, the higher its probability of being masked. This means that the mask probability is dynamic. We also set a complexity threshold of $T$. If the complexity $c$ of a text span exceeds $T$, we will not mask this span. In our experiments, we set $T$ to 0.25 as an empirical value. Following \citet{lewis2020bart}, we set the max mask probability to 0.15, and the length of a text span obeys a Poisson distribution ($\lambda$ = 3). Finally, the mask probability $m$ is calculated as:

\begin{equation}
\label{eq1}
m=\left\{
\begin{aligned}
& 0.15\times (1-\frac{1}{T}\cdot c) , & c \leq T \\
& 0 , & c > T
\end{aligned}
\right.
\end{equation}

The function to mask the text span is denoted as $g(\cdot)$. Given a sentence $x$, the pre-trained model will learn to reconstruct $x$ from the masked sentence:

\begin{equation}
\label{eq2}
l(x) = -logP(x|g(x))
\end{equation}

\subsection{Replacing Complex Words in Ordinary Texts}

We also expect the pre-trained model to learn helpful information from ordinary texts. However, ordinary texts contain more complex words than simple ones, making the pre-trained model learn to reconstruct simple words much less frequently. We introduce the dictionary SimplePPDB++ \cite{maddela2018word} to address this issue. It contains millions of paraphrase rules with readability scores. Therefore, we can replace the complex words in ordinary texts with simple words. Then, the pre-trained model will learn to reconstruct these simple words as in Eq.(\ref{eq2}).

Nevertheless, a word may have different meanings in different sentences. Using a dictionary to replace complex words may change the meaning of the original sentence. Therefore, we use BERTScore \cite{zhang2019bertscore} to calculate the similarity between the original and replaced sentences to avoid this problem. We will discard the replaced sentences if the calculated BERTScore is lower than a similarity threshold. In our experiments, the similarity threshold is set to 0.95 as an empirical value.

\section{Experimental Settings}

\subsection{Continued Pre-training}

We select the BART-Large model to continue pre-training. It is a representative seq2seq model suitable for three main simplification tasks.
We follow the task-adaptive pre-training method \cite{gururangan2020don} and continue pre-training on the training set of the corresponding simplification task, ensuring that the continued pre-training texts have no intersection with the test set.
We refer to the pre-trained models obtained by our strategy collectively as SimpleBART.

\subsection{Simplification Tasks}

We select three representative tasks for experiments: sentence simplification, document-level simplification, and lexical simplification. For sentence simplification, we conduct experiments on Wikiauto \cite{jiang2020neural} and Newsela \cite{xu2015problems}. Wikiauto is only a training set, so we use Turkcorpus \cite{xu2016optimizing} as its validation and test set. Following \citet{sun2023exploiting}, we use SARI \cite{xu2016optimizing} and BERTScore \cite{zhang2019bertscore} as the evaluation metrics. BLEU and FKGL have been proven to be unsuitable for evaluating simplification \cite{sulem2018bleu, tanprasert2021flesch}. For document-level simplification, we conduct experiments on the D-Wikipedia dataset \cite{sun2021document}. We use D-SARI \cite{sun2021document} as the evaluation metric. For lexical simplification, we conduct experiments on LexMTurk \cite{horn2014learning} and BenchLS \cite{paetzold2016benchmarking}. We use precision, recall, and F1 score as the evaluation metrics. For more hyper-parameter setting details, please refer to Appendix \ref{sec:parameters}.

\section{Results}
\subsection{Sentence Simplification}

\noindent
To demonstrate the advantages of our strategy, we develop BART-CP for a fair comparison. It continues pre-training with the same number of steps on the same data using the previous pre-training strategy from \citet{lewis2020bart}. In the continued pre-training stage, text spans are masked randomly.

\begin{table}[h]
\centering
\resizebox{7.5cm}{!}{%
\begin{tabular}{lccccc}
\hline
\multicolumn{1}{l|}{Turkcorpus}   & SARI$\uparrow$ & Keep & Del  & \multicolumn{1}{c|}{Add} & BS$\uparrow$ \\ \hline
\multicolumn{1}{l|}{EditNTS}       & 37.9 & 67.3 & 43.1 & \multicolumn{1}{c|}{3.4} & 0.950    \\
\multicolumn{1}{l|}{T5}       & 37.8 & \textbf{73.5} & 35.6 & \multicolumn{1}{c|}{4.2} & \textbf{0.982}     \\
\multicolumn{1}{l|}{ControlTS} &  \textbf{40.4} & 70.4 & 44.5 & \multicolumn{1}{c|}{6.2} & 0.959    \\
\multicolumn{1}{l|}{BART}       & 38.3 & 65.4 & 44.0 & \multicolumn{1}{c|}{5.6} & 0.973          \\
\multicolumn{1}{l|}{BART-CP}    & 38.6 & 64.6 & 45.9 & \multicolumn{1}{c|}{5.4} & 0.967          \\
\multicolumn{1}{l|}{SimpleBART} & 39.5 & 64.6 & \textbf{47.2} & \multicolumn{1}{c|}{\textbf{6.6}} & 0.972          \\ \hline
                                &      &      &      &                          &           \\ \hline
\multicolumn{1}{l|}{Newsela}    & SARI$\uparrow$ & Keep & Del  & \multicolumn{1}{c|}{Add} & BS$\uparrow$ \\ \hline
\multicolumn{1}{l|}{EditNTS}       & 37.1 & 34.9 & 74.8 & \multicolumn{1}{c|}{1.6} & 0.897     \\
\multicolumn{1}{l|}{T5}       & 36.0 & \textbf{41.8} & 61.9 & \multicolumn{1}{c|}{4.4} & 0.905     \\
\multicolumn{1}{l|}{ControlTS} & 39.7 & 37.6 & 77.3 & \multicolumn{1}{c|}{4.1} & 0.894    \\
\multicolumn{1}{l|}{BART}       & 40.1 & 40.5 & 73.8 & \multicolumn{1}{c|}{6.2} & 0.904     \\
\multicolumn{1}{l|}{BART-CP}    & 40.3 & 41.7 & 72.6 & \multicolumn{1}{c|}{\textbf{6.9}} & \textbf{0.908}     \\
\multicolumn{1}{l|}{SimpleBART} & \textbf{41.6} & 40.5 & \textbf{77.4} & \multicolumn{1}{c|}{\textbf{6.9}} & 0.902     \\ \hline
\end{tabular}%
}
\caption{Results on the Turkcorpus test set and the Newsela test set. We use \textbf{bold} to indicate the best result.}
\label{tab:automatic-evaluation}
\end{table}

We choose EditNTS \cite{dong2019editnts}, T5-base \cite{raffel2020exploring}, and ControlTS \cite{maddela2021controllable} as baselines. T5-base is close to SimpleBART in size. ControlTS achieves the state-of-the-art result on the Newsela dataset. Following \citet{alva2021suitability}, BERTScore$_{precision}$ (BS) is also reported. From Table \ref{tab:automatic-evaluation}, the BS scores of all outputs are high enough, which means that the outputs are of high quality. According to SARI, the most important automatic evaluation metric for sentence simplification, SimpleBART improves SARI values over BART by 1.2 points and 1.5 points, respectively. Overall, it achieves comparable results to the advanced model for the sentence simplification task. We also notice that SimpleBART outperforms BART-CP, demonstrating the effectiveness of our proposed strategy. The example outputs are given in Appendix \ref{sec:example-outputs}. 



\subsection{Lexical Simplification}

\noindent
We focus on generating suitable words using the pre-trained model, which is a critical step in lexical simplification. We follow \citet{qiang2020lexical} and
let the pre-trained models generate several candidate words. BenchLS and LexMTurk are just two test sets, so we continue pre-training on the Wikiauto training set. We choose Paetzold-NE \cite{paetzold2017lexical} and LSBert \cite{qiang2020lsbert} as two baselines. LSBert achieves the state-of-the-art result in this task.
\begin{table}[h!]
\centering
\resizebox{6cm}{!}{%
\begin{tabular}{lccc}
\hline
\multicolumn{1}{l|}{BenchLS}  & F1$\uparrow$   & Precision            & Recall                               \\ \hline
\multicolumn{1}{l|}{Paetzold-NE}   & 23.6    & 27.0                 & 20.9                                \\
\multicolumn{1}{l|}{LSBert}  & \textbf{28.1}     & 24.4                 & \textbf{33.1}                               \\
\multicolumn{1}{l|}{BART}   & 19.2    & 19.6                 & 18.9                               \\
\multicolumn{1}{l|}{BART-CP}  & 25.8  & 26.0                 & 25.7                               \\
\multicolumn{1}{l|}{SimpleBART}  & 27.8 & \textbf{28.0}                 & 27.6                                 \\ \hline
                                & \multicolumn{1}{l}{} & \multicolumn{1}{l}{} & \multicolumn{1}{l}{} \\ \hline
\multicolumn{1}{l|}{LexMTurk} & F1$\uparrow$  & Precision            & Recall                              \\ \hline
\multicolumn{1}{l|}{Paetzold-NE}  & 19.5     & \textbf{31.0}                 & 14.2                                  \\
\multicolumn{1}{l|}{LSBert}   & 26.8     & 30.6                 & 23.8               \\
\multicolumn{1}{l|}{BART}   & 18.8    & 19.2                 & 18.3                   \\
\multicolumn{1}{l|}{BART-CP}   & 26.9   & 27.2                 & 26.6                  \\
\multicolumn{1}{l|}{SimpleBART} & \textbf{28.5}  & 28.7     & \textbf{28.2}         \\ \hline
\end{tabular}%
}
\caption{Results on the BenchLS test set and the LexMTurk test set.}
\label{tab:LS}
\end{table}

As shown in Table \ref{tab:LS}, SimpleBART improves the F1 scores over BART by 8.6 points and 9.7 points, respectively. It achieves comparable results to LSBert. The results also demonstrate that BART needs to gain the ability to generate simple words and the importance of introducing continued pre-training when training data is scarce.

\subsection{Document-level Simplification}

\begin{table}[h!]
\centering
\resizebox{7.5cm}{!}{%
\begin{tabular}{l|cccc}
\hline
D-Wikipedia & D-SARI$\uparrow$ & $D_{keep}$ & $D_{del}$ & $D_{add}$ \\ \hline
BertSumextabs & 39.88  & 35.71 & \textbf{72.06} & 11.87  \\
BART        & 39.84  & 35.87 & 70.26 & 13.40  \\
BART-CP     & 40.13  & 36.21 & 71.54 & 12.64  \\
SimpleBART  & \textbf{41.64}  & \textbf{37.91} & 71.96 & \textbf{15.04}  \\ \hline
\end{tabular}%
}
\caption{Results on the D-Wikipedia test set}
\label{tab:d-wikipedia}
\end{table}
\noindent
SimpleBART also performs well on the document-level simplification task. We choose BertSumextabs \cite{liu2019text}, which achieves the state-of-the-art result on this task as a baseline. Compared with BART, SimpleBART improves the D-SARI value by 1.8 points, making it the new state-of-the-art result.

\section{Analysis}

\subsection{Human Evaluation}

We hired three workers to conduct a human evaluation of the 100 randomly selected outputs of the sentence simplification task. 
Following \citet{dong2019editnts}, workers rate on simplicity (Simp), fluency (Flu), and adequacy (Ade) on a 5-point Likert scale. Following \citet{xu2016optimizing}, we use simplicity gain (S+) to demonstrate how many word-level simplifications occur in sentence simplification.

\begin{table}[h!]
\centering
\resizebox{6cm}{!}{%
\begin{tabular}{l|cccc}
\hline
           & Simp$\uparrow$ & Flu$\uparrow$ & Ade$\uparrow$ & S+$\uparrow$ \\ \hline
EditNTS  & 3.30$^*$ & 4.65$^*$ & 3.56$^*$ & 0.14$^*$   \\ 
T5  & 3.16$^*$ & \textbf{4.91}$^*$ & \textbf{4.47}$^*$ & 0.25$^*$   \\ 
ControlTS  & 3.39$^*$  & 4.67$^*$  & 4.26$^*$  & \textbf{0.60}   \\           
BART       & 3.22$^*$  & 4.80  & 4.31$^*$  & 0.34$^*$   \\
BART-CP    & 3.45$^*$  & 4.68$^*$  & 3.95  & 0.37$^*$   \\
SimpleBART & \textbf{3.62}  & 4.82 & 4.01  & 0.55   \\ \hline
Reference & 3.74 & 4.85 & 4.03 & 0.93$^*$   \\ \hline
\end{tabular}%
}
\caption{Results of the human evaluation. The results significantly different from those of SimpleBART are marked as $^*$ according to the student $t$-test with p<0.05.}
\label{tab:human_evaluation}
\end{table}

Table \ref{tab:human_evaluation} shows that SimpleBART achieves the highest Simp score among all the simplification models, close to that of the reference. SimpleBART also significantly makes more word-level simplifications compared to BART and BART-CP.

\subsection{Domain Adaptation}

Continued pre-training using our strategy on task-related data can improve the results. However, we still want to know if continued pre-training on more data from the same domain and different domains will improve the results. We design the following experiments. 1) Exp1: We continue pre-training on more sentences from Wikipedia and SimpleWiki, except those contained in the Wikiauto dataset. 2) Exp2: We continue pre-training on more sentences in the Newsela corpus, except those contained in the Newsela dataset. The sizes of the above texts used for continued pre-training are roughly five times larger than the simplification training set. 3) Exp3: We continue pre-training on the Newsela training set. 4) Exp4: We continue pre-training on the Wikiauto training set.

\begin{table}[h!]
\centering
\resizebox{6.5cm}{!}{%
\begin{tabular}{l|cccc|c}
\hline
     & SARI$\uparrow$ & Keep & Del & Add & BS$\uparrow$ \\ \hline
Exp1 & 38.9   &  64.9    &  45.7   &  6.0   &  0.968  \\
Exp2 & 41.1   &  39.5    &  77.4   &  6.5   &  0.900  \\
Exp3 & 38.0   &  39.2    &  69.7   &  5.0   &  0.975  \\
Exp4 & 39.6   &  42.1    &  71.1   &  5.7   &  0.907  \\ \hline
\end{tabular}%
}
\caption{Results of domain adaptation experiments. For Exp1 and Exp3, we fine-tuned on Wikiauto and tested on Turkcorpus. For Exp2 and Exp4, we fine-tuned and tested on the Newsela dataset.}
\label{tab:domain-adaptation}
\end{table}

From the results of Exp1 and Exp2 in Table \ref{tab:domain-adaptation}, continued pre-training on more texts from the same domain can still enhance the simplification results. Compared to BART in Table \ref{tab:automatic-evaluation}, the SARI values improve by 0.6 and 1 point, respectively.
From the results of Exp3 and Exp4, continued pre-training on more texts in a different domain can instead harm the results. Compared to BART, the SARI values decrease by 0.3 and 0.5 points, respectively. Thus, we suggest that future researchers use texts within the same domain (e.g., Wikiauto and Wikipedia) for continued pre-training in text simplification.

\subsection{Generating Complex Texts}

There are numerous studies dedicated to simplifying complex texts. Nevertheless, none has attempted to rewrite simple texts into complex ones. We make such an interesting attempt. We have changed our strategy to mask complex words and name the obtained model ComplexBART. When fine-tuning and testing on the Newsela dataset, we use simple texts as input and complex texts as reference.

\begin{table}[h!]
\centering
\resizebox{7.5cm}{!}{%
\begin{tabular}{l|ccccc}
\hline
         & SARI$\uparrow$ & Keep & Del & \multicolumn{1}{c|}{Add} & BS$\uparrow$ \\ \hline
BART     & 35.7 & 53.2 & 50.5 & \multicolumn{1}{c|}{3.3}  & 0.901   \\
ComplexBART & 37.2 & 52.9 & 55.4 & \multicolumn{1}{c|}{3.4}  & 0.900   \\ \hline
\end{tabular}%
}
\caption{Results of generating complex texts.}
\label{tab:hard}
\end{table}

From Table \ref{tab:hard}, ComplexBART improves the SARI value by 1.5 points over the BART model, indicating that the modified strategy can help the pre-trained model learn to generate complex texts. Thus, ComplexBART can serve as a better baseline for generating complex texts in the future.

\section{Comparing SimpleBART with Large Language Models}

Large language models (LLMs) have received widespread attention from researchers recently and have achieved state-of-the-art results on many natural language generation tasks. In this section, we select several representative large models to conduct experiments on text simplification and compare them with SimpleBART. We hope these results can serve as baselines for future research.


We choose those LLMs that provide API or model files to ensure reproducibility. We choose GPT-3.5-Turbo-0301\footnote{\url{https://openai.com/blog/chatgpt}}, FLAN-T5-XL \cite{chung2022scaling}, and LLaMA-7B \cite{touvron2023llama} as LLM baselines and use zero-shot generation. Then, we follow the implementation\footnote{\url{https://github.com/philschmid/deep-learning-pytorch-huggingface/blob/main/training/deepseed-flan-t5-summarization.ipynb}} and fine-tune FLAN-T5-base as another baseline. We collect the training sets of Wikiauto, Newsela, and D-Wikipedia and conduct instruction fine-tuning.

\subsection{Comparison and Analysis}

The comparison of SimpleBART results with those of the LLMs is shown in Tables \ref{tab:compare-ss}, \ref{tab:compare-ds}, and \ref{tab:compare-ls}.

For the sentence-level simplification task, LLaMA and FLAN-T5-XL seem unable to understand the prompt for simplifying sentences, and they are inclined to repeat the original text. However, FLAN-T5-base, only 10\% of the parameters of the above two models, performs better. It illustrates fine-tuning phase can improve performance when the model is not super large. It may be a little strange that GPT-3.5 performs worse than SimpleBART. We find that with the zero-shot setting, GPT-3.5 may not know the ``degree of simplification'' we want. It makes many reasonable changes to the original text, but it also keeps some of the complex parts of the original text.

\begin{table}[h!]
\centering
\resizebox{7.5cm}{!}{
\begin{tabular}{lccccc}
\hline
\multicolumn{1}{l|}{Turkcorpus}          & SARI$\uparrow$                 & Keep                 & Del                  & \multicolumn{1}{c|}{Add}  & BS$\uparrow$                   \\ \hline
\multicolumn{1}{l|}{GPT-3.5}             & 32.4                 & 43.4                 & 43.4                 & \multicolumn{1}{c|}{10.4} & 0.896                \\
\multicolumn{1}{l|}{FLAN-T5}             & 31.5                 & 64.1                 & 29.6                 & \multicolumn{1}{c|}{1.0}  & 0.892                \\
\multicolumn{1}{l|}{LLaMA}               & 29.3                 & 69.3                 & 16.3                 & \multicolumn{1}{c|}{2.3}  & 0.873                \\
\multicolumn{1}{l|}{\begin{tabular}[c]{@{}l@{}}FLAN-T5\\ (Fine-tuned)\end{tabular}} & 36.5                 & 74.4                 & 31.3                 & \multicolumn{1}{c|}{3.8}  & 0.901                \\
\multicolumn{1}{l|}{SimpleBART}          & 39.5                 & 64.6                 & 47.2                 & \multicolumn{1}{c|}{6.6}  & 0.972                \\ \hline
                                         & \multicolumn{1}{l}{} & \multicolumn{1}{l}{} & \multicolumn{1}{l}{} & \multicolumn{1}{l}{}      & \multicolumn{1}{l}{} \\ \hline
\multicolumn{1}{l|}{Newsela}             & SARI$\uparrow$                 & Keep                 & Del                  & \multicolumn{1}{c|}{Add}  & BS$\uparrow$                   \\ \hline
\multicolumn{1}{l|}{GPT-3.5}             & 38.7                 & 32.5                 & 78.1                 & \multicolumn{1}{c|}{5.3}  & 0.897                \\
\multicolumn{1}{l|}{FLAN-T5}             & 32.2                 & 29.7                 & 65.7                 & \multicolumn{1}{c|}{1.3}  & 0.891                \\
\multicolumn{1}{l|}{LLaMA}               & 19.9                 & 35.8                 & 23.2                 & \multicolumn{1}{c|}{0.8}  & 0.822                \\
\multicolumn{1}{l|}{\begin{tabular}[c]{@{}l@{}}FLAN-T5\\ (Fine-tuned)\end{tabular}} & 29.9                 & 40.3                 & 46.7                 & \multicolumn{1}{c|}{2.7}  & 0.902                \\
\multicolumn{1}{l|}{SimpleBART}          & 41.6                 & 40.5                 & 77.4                 & \multicolumn{1}{c|}{6.9}  & 0.902                \\ \hline
\end{tabular}}
\caption{Comparison on the Turkcorpus test set and the Newsela test set.}
\label{tab:compare-ss}
\end{table}

\begin{table}[h!]
\centering
\resizebox{7.5cm}{!}{
\begin{tabular}{l|cccc}
\hline
D-Wikipedia                                                    & D-SARI$\uparrow$ & Keep  & Del   & Add   \\ \hline
GPT-3.5                                                        & 26.68  & 18.45 & 59.36 & 2.25  \\
FLAN-T5                                                        & 26.77  & 15.07 & 64.83 & 0.40  \\
LLaMA                                                          & /      & /     & /     & /     \\
\begin{tabular}[c]{@{}l@{}}FLAN-T5\\ (Fine-tuned)\end{tabular} & 33.22  & 25.08 & 67.50 & 7.09  \\
SimpleBART                                                     & 41.64  & 37.91 & 71.96 & 15.04 \\ \hline
\end{tabular}}
\caption{Comparison on the D-Wikipedia test set.}
\label{tab:compare-ds}
\end{table}

\begin{table}[h!]
\centering
\resizebox{6.5cm}{!}{
\begin{tabular}{lccc}
\hline
\multicolumn{1}{l|}{BenchLS}    & F1$\uparrow$   & Precision & Recall \\ \hline
\multicolumn{1}{l|}{GPT-3.5}    & 36.6 & 36.6      & 36.6   \\
\multicolumn{1}{l|}{SimpleBART} & 27.8 & 28.0      & 27.6   \\ \hline
                                &      &           &        \\ \hline
\multicolumn{1}{l|}{LexMTurk}   & F1$\uparrow$   & Precision & Recall \\ \hline
\multicolumn{1}{l|}{GPT-3.5}    & 31.4 & 31.5      & 31.4   \\
\multicolumn{1}{l|}{SimpleBART} & 28.5 & 28.7      & 28.2   \\ \hline
\end{tabular}}
\caption{Comparison on the BenchLS test set and the LexMTurk test set.}
\label{tab:compare-ls}
\end{table}

For the document-level simplification task, LLaMA over-repeats sentences from the original article, and the generated text is difficult to read. The shortcomings of GPT-3.5 are similar to those of the sentence-level simplification task. Besides, limited by the number of API accesses per minute of OpenAI, we only select 1000 original documents for simplification, which takes nearly five hours.

For the lexical simplification task, neither the LLaMA nor the FLAN-T5 model could understand the instruction to replace complex words with simple words. However, GPT-3.5 outperforms the other models substantially. We also find that GPT-3.5 makes many sensible substitutions not included in the reference, such as replacing ``acquired''with ``earned''. Such results illustrate that LLMs are dominant for this task.

\section{Conclusion}

In this paper, we are committed to adapting the pre-trained model to text simplification. We propose a new pre-training strategy to allow the pre-trained model to learn to generate simple texts. The adapted pre-trained model improves the results on various simplification tasks. 

\section*{Limitations}

The limitation of our method comes from the requirement to identify simple words in simple texts in Section \ref{sec:mask-simple}. The DeepBlueAI we have used is a deep model, meaning it takes much time when inference. In our experiment, it takes 362.78 seconds to identify simple words from 10,000 sentences with an average length of 8.12. We expect that there will be methods with higher identification accuracy and higher inference speed in the future.

Due to page limitations, we have placed the related work in Appendix \ref{sec:related-work} and the ablation experiments in Appendix \ref{sec:ablation}. 

Due to time constraints, we do not perform a human evaluation of the output of LLMs. We hope to conduct a more comprehensive evaluation of the performance of LLMs in the future.



\section*{Ethics Statement}

The texts we have used for continued pre-training come from Wikipedia dumps and the Newsela Corpus. Using Wikipedia dumps requires following the CC-BY-SA license and GFDL. Using Newsela Corpus requires authorization, and we have received it.

This paper contains a human evaluation. We hired three experienced workers to perform it. In the recruiting process, we follow a first-come, first-served order. We paid much more than the local minimum hourly rate.


\section*{Acknowledgements}

We thank Mounica Maddela for her help on the ControlTS baseline. This work was supported by National Key R\&D Program of China (2021YFF0901502), National Science Foundation of China (No. 62161160339), State Key Laboratory of Media Convergence Production Technology and Systems and Key Laboratory of Science, Technology and Standard in Press Industry (Key Laboratory of Intelligent Press Media Technology). Wei Xu and Xiaojun Wan are the corresponding authors.


\bibliography{anthology,custom}
\bibliographystyle{acl_natbib}

\appendix

\section{Related Work}
\label{sec:related-work}

\subsection{Text Simplification}

Text simplification contains sentence simplification, document-level simplification, and lexical simplification.
Sentence simplification is rewriting a complex sentence into a more straightforward and semantically identical sentence \cite{alva2020data}.
Document-level simplification is rewriting an original complex article into a simple article \cite{sun2021document}. Information not relevant to the central meaning can be removed to improve readability. 
Lexical simplification is to replace complex words in a sentence with more straightforward but identical meaning words \cite{paetzold2017survey}. It is usually framed as a pipeline consisting of generating multiple candidate words and developing rules to select the most appropriate word from candidate words.

\subsection{Lexical Complexity Prediction}

The lexical complexity prediction (LCP) task is to assign a value from a continuous scale to represent the complexity of a word \cite{shardlow2020complex}. Given a text and a text span in this text, the model will predict the complexity of this text span. Many studies have been devoted to improving the accuracy of model predictions \cite{gooding2019recursive, paetzold2021utfpr}. On the latest LCP 2021 task \cite{shardlow2021semeval}, the DeepBlueAI model \cite{pan2021deepblueai} achieves state-of-the-art results.

\subsection{Adapting Pre-trained models}

Pre-trained models have been widely used in natural language processing in recent years. However, \citet{gururangan2020don} observe the gap between the language model pre-training domain and the data distribution of the downstream task. 
Since then, researchers have focused on how to adapt pre-trained models to downstream tasks. They have designed new methods for different tasks. Downstream tasks like machine translation \cite{hu2022deep}, sentiment analysis \cite{gu2020train}, and many understanding tasks \cite{yu2022dict} can benefit from the adapted pre-trained models.

\section{Training Parameters}
\label{sec:parameters}

We use the Huggingface transformers \cite{wolf2020transformers} to conduct sentence and lexical simplification experiments. For document-level simplification, we follow \citet{sun2021document} and use Fairseq \cite{ott2019fairseq} to conduct the experiments. We choose the model that performs best on the validation set for testing. The specific parameter settings for each task are shown in Tables \ref{tab:parameter-sentence}, \ref{tab:parameter-lexical}, and \ref{tab:parameter-document}. A detailed description of the dataset sizes is given in Table \ref{tab:datasets}.

Here are the sources of the automatic evaluation methods we use: SARI (\url{https://github.com/mounicam/BiSECT/tree/main/metrics}), BERTScore (\url{https://github.com/Tiiiger/bert_score}), and D-SARI (\url{https://github.com/RLSNLP/Document-level-text-simplification}).

\begin{table}[h!]
\centering
\resizebox{0.48\textwidth}{!}{
\begin{tabular}{rl|rl}
\hline
\textbf{Parameter} & \textbf{Value} & \textbf{Parameter} & \textbf{Value} \\ \hline
epochs             & 10             & max source length  & 128            \\
batchsize          & 64             & max target length  & 128            \\
optimizer          & Adam           & dropout            & 0.1            \\
learning rate      & 5e-5           & weight decay        & 0            \\
warm up steps      & 5000           & seed    & 42             \\ \hline
\end{tabular}}
\caption{Training parameters for sentence simplification.}
\label{tab:parameter-sentence}
\end{table}

\begin{table}[h!]
\centering
\resizebox{0.48\textwidth}{!}{
\begin{tabular}{rl|rl}
\hline
\textbf{Parameter} & \textbf{Value} & \textbf{Parameter} & \textbf{Value} \\ \hline
epochs             & 10             & max source length  & 128            \\
batchsize          & 64             & max target length  & 128            \\
optimizer          & Adam           & dropout            & 0.1            \\
learning rate      & 5e-5           & weight decay        & 0            \\
warm up steps      & 5000           & seed    & 42             \\ \hline
\end{tabular}}
\caption{Training parameters for lexical simplification.}
\label{tab:parameter-lexical}
\end{table}

\begin{table}[h!]
\centering
\resizebox{0.48\textwidth}{!}{
\begin{tabular}{rl|rl}
\hline
\textbf{Parameter} & \textbf{Value} & \textbf{Parameter} & \textbf{Value} \\ \hline
max update steps             & 1e5            & max source length  & 512            \\
max tokens          & 2048             & max target length  & 512            \\
optimizer          & Adam           & dropout            & 0.1            \\
learning rate      & 1e-4           & weight decay        & 1e-4            \\
warm up steps      & 2000           & seed    & 42             \\ \hline
\end{tabular}}
\caption{Training parameters for document-level simplification.}
\label{tab:parameter-document}
\end{table}

\begin{table}[h!]
\centering
\resizebox{5.5cm}{!}{%
\begin{tabular}{lccc}
\hline
\multicolumn{1}{l|}{\textbf{Dataset}}     & \textbf{train}            & \textbf{dev}              & \textbf{test}             \\ \hline
\multicolumn{4}{l}{Sentence simplification}                                               \\ \hline
\multicolumn{1}{l|}{Wikiauto}    & 488K             & \textbackslash{} & \textbackslash{} \\
\multicolumn{1}{l|}{Turkcorpus}  & \textbackslash{} & 2000             & 359              \\
\multicolumn{1}{l|}{Newsela}     & 94K              & 1129             & 1077             \\ \hline
\multicolumn{4}{l}{Lexical simplification}                                                \\ \hline
\multicolumn{1}{l|}{BenchLS}     & \textbackslash{} & \textbackslash{} & 929              \\
\multicolumn{1}{l|}{LexMTurk}    & \textbackslash{} & \textbackslash{} & 500              \\ \hline
\multicolumn{4}{l}{Document-level simplification}                                         \\ \hline
\multicolumn{1}{l|}{D-Wikipedia} & 133K             & 3000             & 8000             \\ \hline
\end{tabular}%
}
\caption{Sizes of the datasets used in experiments.}
\label{tab:datasets}
\end{table}

\section{Ablation Study}
\label{sec:ablation}

\begin{table}[h!]
\centering
\resizebox{7.5cm}{!}{%
\begin{tabular}{l|ccccc}
\hline
           & SARI$\uparrow$ & Keep & Del & \multicolumn{1}{c|}{Add} & BS$\uparrow$ \\ \hline
BART       & 40.1 & 40.5 & 73.8 & \multicolumn{1}{c|}{6.2} & 0.904   \\
BART-S     & 40.9 & 41.6 & 74.2 & \multicolumn{1}{c|}{6.9} & 0.906   \\
BART-T     & 40.9 & 40.6 & 74.9 & \multicolumn{1}{c|}{7.2} & 0.905   \\
SimpleBART & 41.6 & 40.5 & 77.4 & \multicolumn{1}{c|}{6.9} & 0.902   \\ \hline
\end{tabular}%
}
\caption{Results of ablation experiments on the Newsela dataset of the sentence simplification task.}
\label{tab:ablation}
\end{table}

We conduct ablation experiments to explore the different contributions of replacing complex words in ordinary texts (BART-S) and masking simple words in simple texts (BART-T). We continue pre-training and fine-tuning on the Newsela dataset.

From Table \ref{tab:ablation}, both methods in our proposed strategy allow the model to acquire the ability to generate simple words. Their contributions are roughly the same, but the improvement to the SARI value is less than combining them.

\section{Example Outputs}
\label{sec:example-outputs}

\begin{table}[h!]
\centering
\resizebox{7.5cm}{!}{%
\begin{tabular}{l}
\hline
\textbf{Original sentence}                                                                                    \\ \hline
\begin{tabular}[c]{@{}l@{}}gary goddard is the founder of gary goddard\\ entertainment .\end{tabular}         \\ \hline
\textbf{Reference sentence}                                                                                   \\ \hline
gary goddard started gary goddard entertainment .                                                             \\ \hline
\textbf{BART}                                                                                                 \\ \hline
gary is the founder of gary goddard entertainment .                                                           \\ \hline
\textbf{BART-CP}                                                                                              \\ \hline
\begin{tabular}[c]{@{}l@{}}gary goddard is the founder of gary goddard\\ entertainment .\end{tabular}         \\ \hline
\textbf{SimpleBART}                                                                                           \\ \hline
\begin{tabular}[c]{@{}l@{}}gary goddard started a company called gary \\ goddard entertainment .\end{tabular} \\ \hline
\end{tabular}%
}
\caption{In this sentence simplification example, SimpleBART replaces the phrase ``is the founder of" with a simpler phrase ``started a company", which is similar to the reference sentence. Both BART and BART-CP do not simplify the original sentence.}
\label{tab:examples}
\end{table}

\end{document}